\documentclass[times,referee,twocolumn,final,authoryear]{elsarticle}

\usepackage{ycviu}
\usepackage{framed,multirow}

\usepackage{amssymb}
\usepackage{latexsym}

\usepackage{url}
\usepackage{xcolor}
\usepackage{times}
\usepackage{multicol}
\usepackage{color}
\usepackage{xspace}
\usepackage{enumerate}
\usepackage{mathtools}
\usepackage{graphicx}
\usepackage{amsmath}
\usepackage{amssymb}
\usepackage{booktabs}
\usepackage{xcolor}
\usepackage[ruled]{algorithm2e}
\usepackage{algpseudocode}
\usepackage{multirow}
\usepackage{epsfig}
\usepackage{subfigure} 
\usepackage{array}
\usepackage{dsfont}
\usepackage{tabularx}
\usepackage{threeparttable}
\usepackage{ragged2e}
\usepackage{colortbl}
\usepackage{changes}
\usepackage{hyperref} % 用于超链接和PDF元数据的宏包  
\usepackage{eucal}
\usepackage{url}

\usepackage{xcolor}
\newcommand{\RED}[1]{\textcolor{black}{#1}}
\definecolor{newcolor}{rgb}{.8,.349,.1}

\journal{Computer Vision and Image Understanding}

\begin{document}

\begin{frontmatter}

\title{Region-Aware Image-Based Human Action Retrieval with Transformers}

\author[1]{Hongsong Wang}  % \snm{
\ead{hongsongwang@seu.edu.cn}
\author[2]{Jianhua Zhao} 
\author[2]{Jie Gui \corref{cor1}}
\ead{guijie@seu.edu.cn}
\cortext[cor1]{Corresponding author. }

\address[1]{Department of Computer Science and Engineering, Southeast University, Key Laboratory of New Generation Artificial Intelligence Technology and Its Interdisciplinary Applications Nanjing, China}
\address[2]{School of Cyber Science and Engineering, Southeast University, Purple Mountain Laboratories, Engineering Research Center of Blockchain Application, Supervision And Management Nanjing, China}

\received{1 May 2013}
\finalform{10 May 2013}
\accepted{13 May 2013}
\availableonline{15 May 2013}
\communicated{S. Sarkar}

\begin{abstract}
Human action understanding is a fundamental and challenging task in computer vision. Although there exists tremendous research on this area, most works focus on action recognition, while action retrieval has received less attention. In this paper, we focus on the neglected but important task of image-based action retrieval which aims to find images that depict the same action as a query image. We establish benchmarks for this task and set up important baseline methods for fair comparison. We present a Transformer-based model that learns rich action representations from three aspects: the anchored person, contextual regions, and the global image. A fusion transformer is designed to model the relationships among different features and effectively \RED{fuse} them into an action representation. Experiments on \RED{both} the Stanford-40 and PASCAL VOC 2012 Action datasets show that the proposed method significantly outperforms previous approaches for image-based action retrieval.
\end{abstract}


\end{frontmatter}

%\linenumbers


% \section*{Acknowledgments}

\section{Introduction}
Understanding human actions from visual data is a crucial task in the computer vision community, with tremendous applications in video surveillance, autonomous driving, human–robot interaction, health monitoring, etc. Action recognition and action retrieval are two important tasks for human action understanding. \RED{The former} aims to classify human action in a given video or image, while \RED{the latter} strives to find videos or images that depict the same action as a query video or image.

Despite great progress has been made, most works focus on video-based human action understanding and particularly on the task of video-based action recognition \cite{kong2022human}. Action retrieval has received less attention than action recognition in the academic community. Limited existing works of human action retrieval are primarily targeted at actions in the video \cite{wray2019fine, kico2022towards}. Although videos contain more information that serves better action recognition, understanding actions in videos requires more computational resources.

\begin{figure}
	\includegraphics[width=0.95\linewidth]{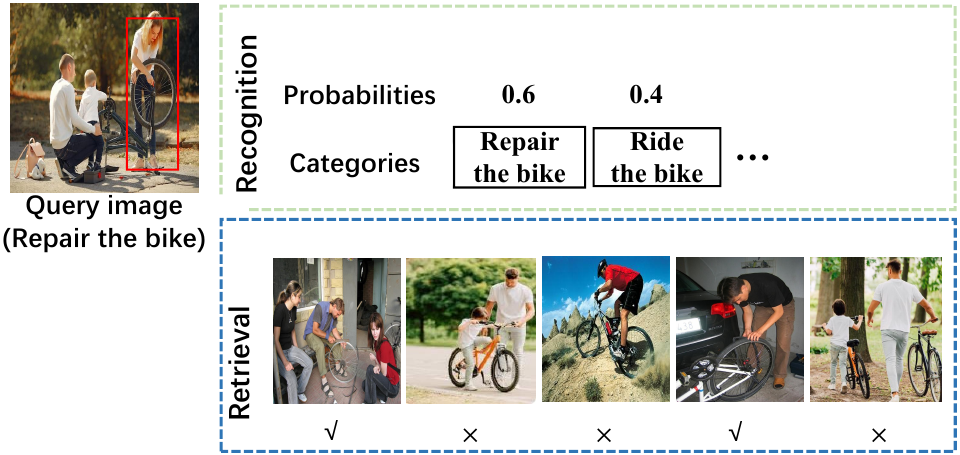}
	\caption{The \RED{illustration} of image-based human action retrieval. The red box indicates the anchored region of the person who performs the action.}
	\label{fig:problem}
\end{figure}
The understanding of human actions in images, which lack temporal information, is a challenging task since the semantic information contained in images is sparser and more abstract than that in videos.
Image-based human action recognition aims to distinguish the static action categories in images, usually the bound box of the person who performs the target action is provided to recognize the corresponding action.
Previous approaches can be roughly categorized into body part-based \cite{zhao2017single, li2020recognizing} and human-object interaction-based \cite{wu2021improved, girish2020understanding}. The part-based approaches require semantic labeling of body parts to allow precise analysis of human actions, and pose additional challenges in terms of annotation efforts and computational complexity. The human-object interaction-based approaches require additional annotation of objects and are limited in their ability to handle actions involving a person manipulating an object. The existing works on image-based human action recognition primarily focus on closed-set scenarios, while research on open-set human action understanding remains scarce.

Image-based visual retrieval aims to search and retrieve relevant images by providing a sample image describing visual features. Typical tasks involves content-based image retrieval~\cite{dubey2021decade}, person re-identification~\cite{ye2021deep,zhang2023graph}, sketch-based image retrieval~\cite{bhunia2020sketch} and image-based patent retrieval~\cite{wang2023learning}. However, these works primarily concentrate on retrieving contents such as objects or landmarks within images, whereas retrieving human actions from images remains an underexplored problem.

Although image-based action retrieval can be seen as a particular application of instance-level image retrieval, it is more challenging than retrieving the landmark or product image. This is due to the fact that understanding human actions involves a semantically higher level of perception than object recognition, and there are often more significant appearance differences among images depicting the same action. 
Fig.~\ref{fig:problem} shows the example of image-based action retrieval. Considering the action of ``repairing the bike," the query image can easily be confused with the action of ``riding the bike," particularly when the appearances or persons depicted in the image are very similar. 
\RED{In addition, image-based human action retrieval not only promotes deeper research into visual human action understanding, but also demonstrates significant practical potential in real-time and efficient action recognition/retrieval scenarios.}
In this paper, we study the neglected but important task of image-based action retrieval. As the image may contain multiple persons, an anchored region is required to indicate the person who performs the action. Since the action involves interactions between the person and contextual objects, we hypothesize that the possible regions located close to the human in the image also play a more important role in understanding human actions. \RED{Considering that the Transformer \cite{2017Attention} is capable of learning a joint representation that integrates information from multiple modalities, rendering it highly suitable for fusion tasks, we employ the Transformer to fuse the representations of the person, contextual objects, and their interactions. In the query example of repairing the bike in Fig.~\ref{fig:problem}, simple feature fusion strategies like averaging pooling or concatenation are unable to capture intricate interactions and relationships between the person and related objects. In contrast, Transformer can learn the joint representations that encapsulates the complementary information between them.}
	
% We hypothesis that the features located close to the human body and located on the body in the image play a more important role in understanding human actions.
% In order to better solve this problem, a more powerful learner should be proposed to better understand human actions in static images for human action recognition and retrieval. In this paper, we try to propose a method for more effective retrieval of human actions in static images.
% Human actions have the property that human actions are always closely related to the human body, as well as to some background in the vicinity of the body. It indicates that the features near the body in the image are more important for understanding the body action, compared to the features remote from the body, which are less important for understanding the action. That doesn't mean that features from the body part are not important, for example, the features nearby the body part in action ``playing violin'' and ``playing guitar'' may provide less role in understanding.
% Therefore, we propose that the features close to the human body, not only those from the body part, are more important for understanding human actions. 

Accordingly, we propose an end-to-end model that learns multi-level representations of different granularities for image-based human action retrieval. Multi-level representations are representation of anchored person, representations of contextual regions and the global representation. The representation of anchored person and the global representation capture the discriminative information of the human performing the action and the whole image, respectively. To obtain reliable contextual regions, we first employ proposal generation to generate bounding boxes encompassing potential object regions, then a simple contextual regions ranking module is designed to select proposals. Finally, a fusion transformer module is proposed to fuses features from different levels into an efficient representation using type embedding and positional embedding.

In summary, the contributions of this paper are as follows:
\begin{itemize}
	\item We empirically study the neglected task of image-based human action retrieval, and establish new benchmarks and important baselines to promote research in this field.
	\item \RED{We introduce an efficient Region-aware Image-based human Action Retrieval with Transformers (RIART), which leverages both person-related and contextual object cues, and employs a fusion transformer module for human action retrieval.}
	% To fully exploit person-related and contextual cues for action retrieval, we present an end-to-end model that learns multi-level representations.
	% \item We design a fusion transformer module to fuse different features and obtain an efficient action representation.
	\item The proposed method has significant performance improvement than other recently proposed instance-level image retrieval methods on several important benchmarks.
\end{itemize}

\section{Related Work}
\subsection{Image-Based Action Recognition}
Image-based action recognition is a complex and challenging endeavor that requires a deep comprehension of the human actions depicted within the image. Deep learning has been widely applied to this task and achieved remarkable results. We categorize recent works on this topic into three groups: body part-based method, human-object interaction-based method and transfer learning-based method.

Body part-based method focuses on extracting features from different parts of the human body and learning their semantic meanings. For instance, \cite{zhao2017single} divide the human body into several parts and assign each part a semantic part action label. \cite{li2020recognizing} propose a body structure exploration sub-network that captures the global and local structure information of human bodies. These methods aim to exploit the fine-grained details of human poses for action recognition.

Human-object interaction-based method studies the relationship between humans and objects in the scene and use it as a cue for action recognition. For example, \cite{wu2021improved} \RED{introduce an efficient relation module that integrates both human-object and scene-object relationships for the purpose of enhancing action recognition capabilities.} \cite{girish2020understanding} develop an action recognition method based on target detection and pose estimation that leverages the spatial relationship between humans and objects and the hierarchical relationship between action categories. These methods aim to exploit the contextual information of objects for action recognition.

Transfer learning-based method uses pre-trained models on large-scale datasets and fine-tune them on specific action recognition datasets. For example, \cite{mohammadi2019ensembles} employ the transfer learning technique to address the problem of data scarcity in action recognition datasets. \cite{chakraborty2021transfer} use a pre-trained model on ImageNet and fine-tune it on a small-scale action recognition dataset using a novel loss function. These methods aim to exploit the generalization ability of pre-trained models for action recognition.\RED{ \cite{8578655} aim to hallucinate temporal characteristics from static images, leveraging video memory techniques, in order to enhance action recognition capabilities with few still images. }

\RED{To the best of our knowledge, only a very limited number of works~\cite{ramanathan2015learning,li2011actions} have studied image-based action retrieval, primarily owing to the scarcity of benchmarks. The work in \cite{ramanathan2015learning} follows the traditional Bag-of-Words pipeline, which is based on manually designed local feature descriptors. The work \cite{li2011actions} aims to learn global image-level action representations by leveraging the semantic relationship between different actions. However, both models are only suitable for simple images that contain a single action and are not scalable enough for large-scale applications. In contrast, we comprehensively study image-based action retrieval and introduce an efficient, deep learning-based pipeline.
}

% As one of the few works of image-based action retrieval, Ramanathan \textit{et al.} \cite{ramanathan2015learning} use language cues, visual cues, and logical consistency to match action images with textual descriptions. 
% Some works try to not use the human box as an input for action recognition. For example, Feng \textit{et al.} \cite{feng2017boxless} propose a method that uses a recurrent neural network model of visual attention to extract information from a sequence of fixations. Liu \textit{et al.} \cite{liu2019loss} introduce a human-mask loss to automatically guide the activations of the feature maps to the target human who is performing the action. These methods aim to avoid the dependency on human detection for action recognition.

\subsection{Human Action Retrieval}
Content-based video retrieval~\cite{spolaor2020systematic} aims to search for relevant videos based on a query video. Given that videos contain vast amounts of raw data, techniques such as video segmentation, feature extraction, and dimensionality reduction can be employed to enhance efficiency. The challenge lies in achieving high accuracy while ensuring efficient computation and low storage consumption~\cite{jiang2021learning}.

Human action retrieval is a new and challenging topic that aims to find videos or images that contain specific human actions. \RED{Early works rely on efficient feature matching methods based on local features~\cite{2013Human}.}
Most existing methods retrieve human actions from videos by describing local features of the motion patterns \cite{veinidis2019effective}. 
% Ramezani \textit{et al.} \cite{ramezani2018motion} introduce a method based on fractal dimension to measure the complexity of motion patterns on different scales and use it as a feature for human action representation and retrieval. 
\cite{kico2022towards} present a metric learning-based method to learn a feature space that preserves the similarity of human actions and enables efficient content-based retrieval. 
Some other methods \cite{veinidis2019effective} use unsupervised approaches to cluster and retrieve action videos based on motion features. \RED{There are also works~\cite{10.1007/978-3-031-12423-5_18} that focus on content-based action retrieval from human skeleton sequences. 
It should be noted that human action retrieval is also closely related to zero-shot action recognition~\cite{2021Zero}, which aims to recognize novel action categories without training data. However, zero-shot action recognition emphasizes generalization to unseen classes, while action retrieval emphasizes accuracy and efficiency of retrieval.
}

\subsection{Instance-level Image Retrieval}
\RED{Instance-level image retrieval involves searching a vast database to retrieve images that contain an object identical to the one specified in a query image.} This task requires effective and efficient methods to represent and compare images. Some works \cite{simeoni2019local, cao2020unifying} adopt a geometric verification approach to rerank images based on local features. Recently, some methods \cite{radenovic2018fine, tan2021instance, wu2022learning} have aimed at learning more powerful representations of images using global features.

% Philbin \textit{et al.} \cite{philbin2007object} propose a method that combines local invariant features with a visual vocabulary and a spatial verification step to retrieve images of specific objects. 
\cite{simeoni2019local} propose a new Deep Spatial Matching (DSM) method for image retrieval that uses an initial ordering of image descriptors extracted from the activation layer of a convolutional neural network, followed by a set of local features approximating the same sparse 3D activation tensor. 
\cite{cao2020unifying} \RED{integrate global and local features within a unified deep model, facilitating precise retrieval through efficient feature extraction. This integration is achieved by leveraging generalized mean pooling for capturing holistic global characteristics and attentive selection mechanisms for pinpointing salient local details.}
% Arandjelovic \textit{et al.} \cite{arandjelovic2016netvlad} introduce an end-to-end method with a new generalized VLAD layer that aggregates convolutional features into a compact global descriptor. They also develop a training procedure based on a new weakly supervised ranking loss.
\cite{radenovic2018fine} fine-tune CNNs for image retrieval on extensive collections of unordered images. They use hard negative mining and feature normalization to improve the performance of global descriptors.
\cite{tan2021instance} introduce Reranking Transformers as a versatile model that incorporates both local and global features to refine the ranking of matched images in a supervised manner, thereby streamlining the typically resource-intensive process of geometric verification. 
\cite{wu2022learning} propose a tokenizer module to aggregate deep local features of images into a few visual tokens, then enhance these tokens by a refinement block, and finally generate the global representation by concatenating these visual tokens. 

These methods effectively retrieve landmarks and products, yet they are not fit for retrieving action images which contain richer semantic information than landmarks. Our approach retrieves action images more efficiently by better understanding the semantics in images.
% \section{Image-Based Human Action Retrieval}
% \subsection{Problem Statement}
% \subsection{Evaluation Metrics}

\section{Method}
\begin{figure*}
	\includegraphics[width=0.95\textwidth]{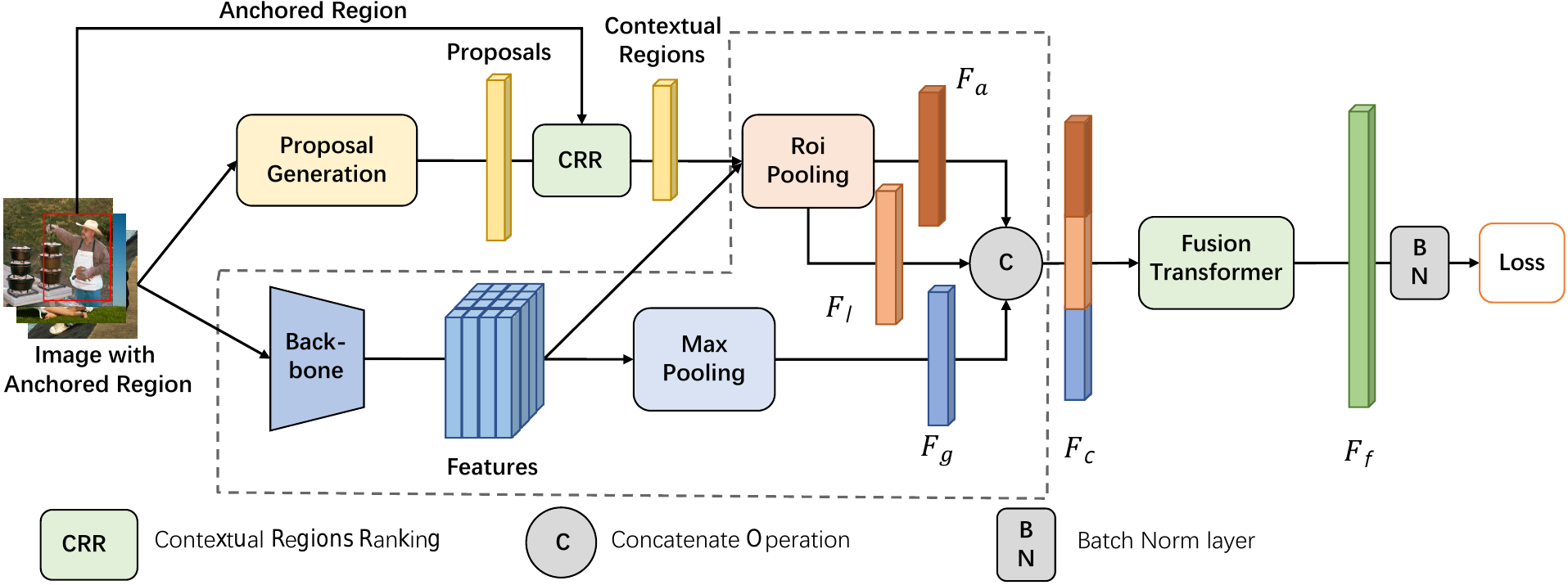}
	\caption{\RED{The overall architecture of the proposed Region-aware Image-based human Action Retrieval with Transformer (RIART). The RIART takes an image and the corresponding anchored region of the human bounding box as input, generating a human action representation that can be utilized for action retrieval or recognition. To acquire contextual regions associated with the action, an off-the-shelf proposal generation module is employed to generate object proposals, which are then filtered and ranked by a contextual regions ranking module. Multi-level action representations are obtained, specifically, the representation of the anchored person $F_a$, representations of contextual regions $\mathbf{F_i}$, and the global image-level representation $F_g$. Subsequently, these representations are fused by a fusion transformer module to yield the final representation $F_f$. 
 }}
	\label{fig:overall_arc}
\end{figure*}

To enhance understanding of human actions within images, we aim to harness representations that span various levels of granularity: the anchored person, contextual regions, and the entire image. The anchored person specifically denotes the human body, whereas contextual regions represent objects proximate to the human that are pertinent to the action being depicted. \RED{We propose an efficient Region-aware Image-based human Action Retrieval with Transformer (RIART). The overall architecture of the proposed method is illustrated in Fig. \ref{fig:overall_arc}, and the details of its key components are described as follows.}

% backbone, swintranformers, multi-level representations,
\subsection{\RED{Problem Formulation}}

\begin{figure}[tbp]
	\subfigure[]{
		\label{fig:query:a}
		\includegraphics[width=1.58in]{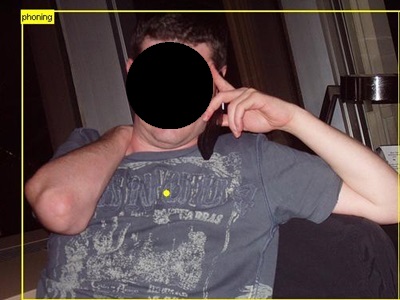}}
	\hspace{0.00in}
	\subfigure[]{
		\label{fig:query:b}
		\includegraphics[width=1.58in]{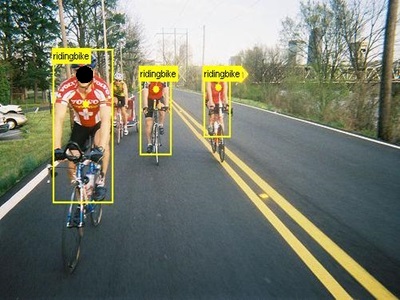}}
	\caption{\RED{Examples query image for action retrieval. Human bounding boxes indicate the person performing a certain action. (a) Simple image that contains only one human action. (b) Complex image that contains multiple actions from different persons. Note that actions of different persons in the same image can be different.}}
	% module on the Stanford-40 dataset
\label{fig:query}
\end{figure}
\RED{Image-based human action retrieval refers to the process of retrieving images from a database that depict specific human actions. Unlike video-based human action retrieval, the actions of interest for this task are typically static actions that can be exemplified in a single image. Examples of query image are shown in Fig. \ref{fig:query}. The query image may contain multiple action classes within the same image. We assume that a given human bounding box in an image only encompasses one action class. If the query image depicts multiple persons performing different actions, each image with distinct human bounding boxes is regarded as a separate query example.}

\subsection{Representation of Anchored Person}
The input image may contain multiple persons who perform different actions. To understand the action performed by a particular person in the image, an anchored person should be used to identify the position of the human. 
% We use both two kinds of anchored region: human annotation and .

To facilitate image-based action understanding, it is crucial to learn the representation of the anchored person. To achieve this, we utilize ROI pooling to extract this representation based on the provided bounding box. Let $\mathbf{F}$ be the output feature map of the backbone network, $B_a$ be the bounding box of the person, representation of the anchored person $F_a$ is
\begin{equation}
	{F_a} = \mathrm{ROI\_Pooling}(\mathbf{F}, B_a).
\end{equation}
\RED{where $\mathrm{ROI\_Pooling}(\cdot)$ represents the Region of Interest Pooling function~\cite{ren2015faster}, which takes as input a feature map from and a bounding box defining a Region of Interest (ROI), and outputs a fixed-size feature map from the specified ROI.}

\subsection{Representations of Contextual Regions}
Many human actions involve interactions with surrounding objects and environments, therefore, learning representations of contextual regions around the individual performing the action is crucial for action understanding. We employ a proposal generation module to generate bounding boxes encompassing potential object regions. The proposal generation process often generates thousands of proposals, most of which are unrelated to the concerned action. Therefore, we design a contextual regions ranking module to select a few useful proposals of contextual regions.

The contextual regions ranking module selects the proposals nearest to the anchored person, with confidence scores exceeding a predefined threshold. The Intersection-over-Union (IoU) metric is employed to measure the similarity between two boxes. We select the top $k$ bounding boxes with the highest IoU scores with the bounding box of anchored person. 
This process is formulated as
\begin{equation}
	\begin{aligned}
		\widetilde{\mathbb{P}} =&\  \mathrm{Rank\_Select}(\mathrm{IoU}(\mathbb{P} , B_a)), \\
		\mathbf{F_{l}} =&\ \mathrm{ROI\_Pooling}(\mathbf{F}, \widetilde{\mathbb{P}}) ), 
	\end{aligned}
	\label{eq:F_semantic}
\end{equation}
where $\mathbb{P}$ denotes the proposals with the confidence scores above the threshold, $\widetilde{\mathbb{P}}$ denotes the selected top-k proposals about contextual regions, $\mathbf{F_{l}}$ is the set of representations of contextual regions \RED{(for symbols' fonts, italic and bold styles are used to denote feature vectors and sets of features, respectively)}. 

The whole image contains rich contextual information about the action. To utilize the overall contextual information for the whole image, we also use the global representation of the image for action understanding. The global image-level feature is 
\begin{equation}
	F_{g} = \  \mathrm{Max\_Pooling}(\mathbf{F}),
	\label{eq:F_global}
\end{equation}
where $\mathrm{Max\_Pooling}$ denotes the max pooling operation.
% In order to extract features from the image data, which contains low-level semantic information and rich structural information, we employ a CNN network as a feature extraction module. The CNN network has the advantages of spatial invariance and strong adaptability, which make it suitable for modeling the local features of the data. We adopt the ResNet-50 \cite{he2016deep} network, which is pre-trained on the ImageNet \cite{deng2009imagenet} dataset, as the backbone of our multi-level representations learning module.
% We feed an image $I \in \mathbb{R}^{C\times H\times W}$ into the multi-level representations learning module, where $C$ is the number of channels, and $H$, $W$ are the height and width of the image respectively. We obtain the feature map $f\in \mathbb{R}^{2048\times 7 \times7}$ from the conv5 layer of ResNet-50 as the output features of the backbone. This feature map $f$ is then used to generate our proposed features. 

\subsection{Fusion Transformer}
\begin{figure}[h]
	\centering
	\includegraphics[width=0.95\linewidth]{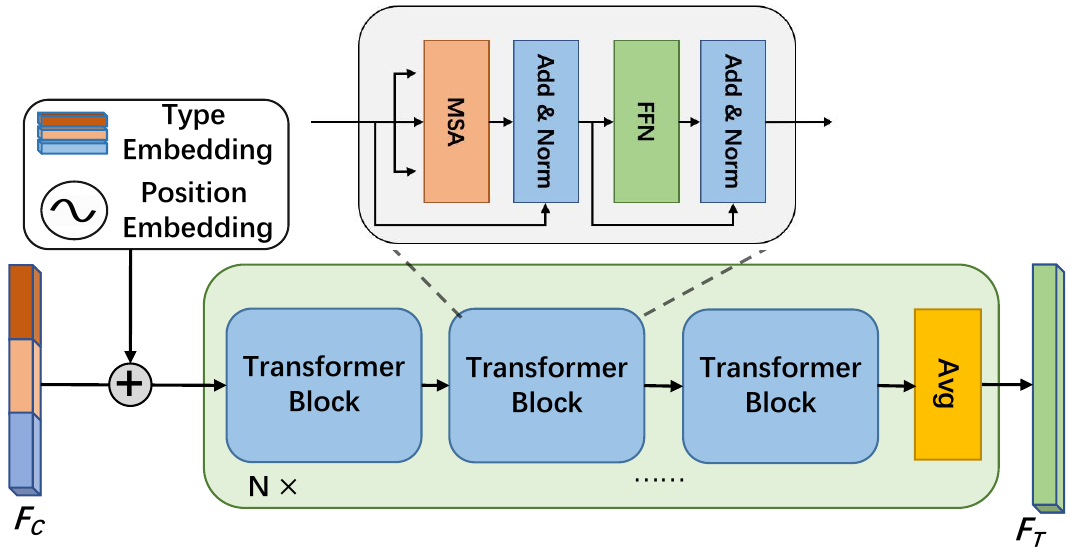}
	\caption{\RED{The architecture of Fusion Transformer module. The input $F_c$ denotes multi-level action representations, i.e., representation of the anchored person $F_a$, representations of contextual regions $\mathbf{F_i}$, and the global image-level representation $F_g$. The Fusion Transformer aims to capture the complementary information among these diverse input representations and output a learned joint representation. To facilitate the learning of this complementary action representation, type embeddings and position embeddings are employed. Similar to the standard Transformer architecture, the Fusion Transformer module consists of $N$ blocks.
	}}
	\label{fig:transformer_arc}
\end{figure}

Considering the advantages of the Transformer in global dependency modeling, we design a Transformer-based fusion module to merge multi-level representations from different aspects. The fusion transformer, shown in Fig.\ref{fig:transformer_arc}, consists of a cascade of $N$ Transformer blocks. The input of Transformer-based fusion is a set of multi-level representations, i.e., $\{F_{a}, F_{g}, \mathbf{F_{l}}\}$. 
We flatten these features and convert them into a sequence of tokens. The number of input tokens is $(k+2)$, where $k$ is the number of selected contextual regions. The dimension of the concatenated features is ${(k+2)\times D}$, where $D$ is the dimension of representation. 

Subsequently, we add positional embedding and type embedding before feeding concatenated features into the transformer block. The type embedding $typ$ identifies different aspects of features in the concatenated sequence of tokens. The positional embedding $pos$ provide information about the position of the feature in the original feature map after being flattened into tokens. 
The calculation of positional embedding is
\begin{equation}  
	pos_t^{(j)}=\left\{\begin{array}{l}
		\sin \left(w_i t\right), \quad \text { if } j=2 i \\
		\cos \left(w_i t\right), \quad \text { if } j=2 i+1
	\end{array}\right.
	\label{eq:pos_embedding}
\end{equation}
where \RED{$t$ represents the position in the sequence of different features, $j$ indexes the dimension of the feature and $w_i$ is a frequency term that varies with the iterating index $i$. The $w_i$ is
}
$$
w_i=\frac{1}{10000^{2 i / D}}, i=0,1, \ldots, \frac{D}{2}-1.
$$
\RED{This formulation captures both fine-grained and coarse-grained positional information across different dimensions of features.}

The output sequence also has $(k+2)$ tokens, and we simply average them to get final features. The above processes could be summarized as follows
\begin{equation}
	\begin{aligned}
		emb_{i} = pos_{i} + typ_{i}, i= (1, 2, 3),& \\
		F_{c} = Concat(F_{a} + emb_{1}, F_{g} + emb_{2}, \mathbf{F_{l}} + &emb_{3}),\\
		F_{T} = \mathrm{Avg}(\mathrm{Fusion\_Transformer}(F_{c})), \\
	\end{aligned}
\end{equation}
where the $pos_{i}$ denotes the positional embedding while the $typ_{i}$ denotes the type embedding, $\mathrm{Avg}$ is the average operation. The output representation $F_{T}$ is used to recognize or retrieve human action in the image.

\RED{It should be noted that Transformer-based fusion has also appeared in recent works \cite{2022Dual} for multi-modal human action recognition. For instance, a cross-modality fusion module is applied to learn better multi-modal representations \cite{2022Dual}. Unlike the fusion modules of these approaches, the proposed Fusion Transformer aims to learn efficient representations from a single image modality by fusing multi-level or multi-grained features. Additionally, type embedding and position embedding are designed to facilitate the feature learning.
} 

\subsection{Training Strategy}
We also add a batch norm layer on top of $F_{T}$ for better performance.
We train our model through the action classification task to learn the representation $F_T$. We use a two layers MLP with ReLU \cite{glorot2011deep} activation function as the classifier to classify human actions in images. The input of the classifier is $F_T$ which is the output feature of the Transformer module. We adopt cross-entropy loss as the loss function of our classifier:
\begin{equation}
	Loss = -\sum_{i=1}^{n}{p(x_i)\log{q(x_i)}},
\end{equation}
where the $p(x_i)$ is the probability of the ground truth label, $q(x_i)$ is the probability of the label predicted by the classifier and n is the number of classes. 

The action classification task is also the pretext task for the action retrieval task. The representation $F_T$ learned by the action classification task is used to represent the action image at retrieval time.

\section{Experiments}
% In this section, we conduct extensive experiments on image action datasets to evaluate our method. We evaluate our method on both classification and retravel tasks.

\subsection{Datasets and Implementation Details}
\noindent \textbf{Stanford-40.} The Stanford 40 Action dataset~\cite{yao2011human} is an image-based static action recognition dataset with a diverse range of 40 actions. Each image is annotated with a bounding box which indicates the person who is performing the specific action. This dataset comprises a total of 9532 images, with each action class containing a balanced and diverse representation of the respective action, ranging from 180 to 300 images. 

\noindent \textbf{PASCAL VOC.} The PASCAL VOC 2012 Action Classification~\cite{everingham2015pascal} dataset comprises a comprehensive collection of annotated images, specifically consisting of 2296 training images and 2292 validation images. There are 10 actions which spans a range of categories. For each image, annotations are provided to indicate the presence of an action and its corresponding class label.

\noindent \textbf{Implementation Details.} For proposal generation, we use FasterRCNN~\cite{ren2015faster} with pre-trained parameters. 
% The backbone of representation learning module is the ResNet-50 pre-trained on ImageNet. 
\RED{The default backbone for the representation learning module is ResNet-50 \cite{he2016deep}. Additionally, a Transformer-based backbone, specifically Swin-T \cite{liu2021swin}, is also employed due to its comparable model size to ResNet-50.}
The fusion transformer has 3 Transformer blocks with 8 heads each and is trained from scratch. The input size of images is $224\times224$, with data augmentation including horizontal flipping, random crop, random color jitter, and mix-up during training. We use the batch size of 128 to train our model on a single Nvidia RTX 3090 GPU. The early-stop strategy is used and the max training epochs is set to 150. AdamW is used to optimize the model with an initial learning rate is 0.00003. 
% An exponential decaying scheduler is adopted to decay the learning rate to 0 gradually. 
The number of selected bounding boxes $k$ is set to 10. The feature dimension $D$ is set to 2048. 
% We set the number $k=10$ of selected proposals. The feature map size $s$ of multi-level features is set to 1 for features of three different aspects, and the dimension $D$ is set to 2048. The number $n$ of tokens in $F_c$  is 12.

\subsection{Experimental Setup}
% The experimental setup on the action retrieval task is different from the task of recognition. 
\noindent \textbf{Image-based Action Retrieval.} As there are no benchmarks for image-based action retrieval, we establish experimental settings and conduct baselines for subsequent analysis and comparison. The experimental settings for image-based action retrieval are constructed based on those of action recognition on \RED{both} the Stanford-40 and PASCAL VOC datasets. Specifically, we train the action representation model on the training set, and consider the validation set as both the query and gallery sets for action retrieval. As the query set and gallery set are the same set, we remove the query sample from the gallery set while retrieving. \RED{The query set refers to the collection of query images that are used as input to the image-based retrieval model, while the gallery set is a collection of images that the image retrieval system searches through to find matches for the query images.}
% We train the model of action retrieval through the action recognition task and use the feature $F_T$ as the representation of the image.

% The action retrieval task is to select an image containing a certain human action as a query and try to find images with the same human action in the whole dataset (excluding the one used as a query). Before performing action image retrieval, we use the action image classification task as a pretext task for action image retrieval. We use the output features $F_T$ of the fusion transformer module as the representation of the images at the retrieval task after the model has been trained by the classification task. 

\noindent \textbf{Baselines.} To demonstrate the effectiveness of our approach, we select four representative baseline methods: Reid strong baseline (Strong baseline) \cite{luo2019bag}, DELG \cite{cao2020unifying}, Reranking Transformer (RRT) \cite{tan2021instance}, and Token \cite{wu2022learning}. 
Reid strong baseline \cite{luo2019bag} is a method for retrieving the same person that appears in photos taken from different cameras. We use this method for the action retrieval task where we set the query images and the gallery images as coming from different cameras and treat the same class of actions as the same person in the original setting of Reid strong baseline.
DELG \cite{cao2020unifying} is an instance-level retrieval method for identifying and retrieving different views of a specific instance in a database. We mainly use the global feature in DELG to conduct the action retrieval task.
Reranking Transformer \cite{tan2021instance} is another instance-level image retrieval method that learns features contrastively between samples using a Transformer.
Token \cite{wu2022learning} is also an instance-level method for image retrieval, it designs a tokenizer module to aggregate different features into a few vision tokens.

\noindent \textbf{Evaluation Metrics.} For action retrieval, we adopt the widely used mAP, Rank-1, Rank-5 as the evaluation metrics. \RED{Reranking~\cite{luo2019bag, zhang2023graph} is a common post-processing step of reordering the initial retrieved results based on additional criteria, with the goal of improving the retrieved results.} Unless otherwise specified,  we use the reranking method in \cite{luo2019bag}, and set the parameters $k1=200$, $k2=20$, and $lambda=0.3$.

% results of original resnet
\begin{table}[tbp]
	\centering
	\caption{Results of human action retrieval on the Stanford-40 dataset. \RED{In order to keep the model lightweight, we employ not only ResNet50 but also Swin-T~\cite{liu2021swin} as the backbone. The choice of Swin-T stems from the fact that its model size is comparable to that of ResNet50. }}
	\resizebox{0.49\textwidth}{!}{ 
		\begin{tabular}{l|ccc|ccc}
			\toprule
			\multicolumn{1}{l|}{\multirow{2}[0]{*}{Method}} & \multicolumn{3}{c|}{w/o reranking} & \multicolumn{3}{c}{w/ reranking} \\
			\noalign{\vskip 0.5mm}
			\cline{2-7}
			\noalign{\vskip 0.8mm}
			& \multicolumn{1}{c}{mAP} & \multicolumn{1}{c}{Rank-1} & \multicolumn{1}{c|}{Rank-5} & \multicolumn{1}{c}{mAP} & \multicolumn{1}{c}{Rank-1} & \multicolumn{1}{c}{Rank-5} \\
			\midrule
			Strong baseline \cite{luo2019bag} & 59.1 & 73.6 & 83.9  &  63.7 & 70.7 & 81.5 \\
			DELG global \cite{cao2020unifying}  & 28.1 & 57.8 & 81.1 &  42.9 & 60.6 & 79.1 \\
			RRT \cite{tan2021instance} & -- & -- & -- & 53.9 & 53.4 & 55.9 \\
			\RED{Swin-T} \cite{liu2021swin} & 78.2 & 85.2 & \textbf{93.1} & 81.1 & 84.3 & \textbf{92.2} \\
			R50-Token \cite{wu2022learning} & 62.9 & 78.7 & 89.2 & 71.8 & 77.8 & 87.1 \\
			\midrule
RIART (ResNet-50) & 73.2 & 83.1 & 91.3 & 77.5 & 81.5 & 90.1 \\
\RED{RIART (Swin-T)} & \textbf{80.7} & \textbf{85.6} & \textbf{92.5} & \textbf{81.9} & \textbf{84.5} & \textbf{91.7} \\
			% Ours (Swin-T) &   &  & \textbf{81.9} & \textbf{84.6} & \textbf{91.8} \\
			\bottomrule
	\end{tabular}}
	\label{tab:retrieval_stanford}
\end{table}

% results of original resnet
\begin{table}[tbp]
	\centering
	\caption{Results of static action retrieval on the PASCAL VOC dataset.}
	\resizebox{0.49\textwidth}{!}{ 
		\begin{tabular}{l|ccc|ccc}
			\toprule
			\multicolumn{1}{l|}{\multirow{2}[0]{*}{Method}} & \multicolumn{3}{c|}{w/o reranking} & \multicolumn{3}{c}{w/ reranking} \\
			\noalign{\vskip 0.5mm}
			\cline{2-7}
			\noalign{\vskip 0.8mm}
			& \multicolumn{1}{c}{mAP} & \multicolumn{1}{c}{Rank-1} & \multicolumn{1}{c|}{Rank-5} & \multicolumn{1}{c}{mAP} & \multicolumn{1}{c}{Rank-1} & \multicolumn{1}{c}{Rank-5} \\
			\midrule
			Strong baseline \cite{luo2019bag} & 26.0 & 51.3 & 82.6 &  32.0 & 49.6 & 78.2 \\
			DELG global \cite{cao2020unifying} & 31.2 & 58.5 & 86.1 & 43.6 & 60.1 & 84.2 \\
			RRT \cite{tan2021instance} & -- & -- & -- & 50.7 & 50.1 & 53.7 \\
			R50-Token \cite{wu2022learning} & 46.3 & 65.5 & 84.5 & 52.2 & 62.9 & 83.4 \\
			\midrule
			RIART (ResNet-50) & \textbf{56.3} & \textbf{73.1} & \textbf{88.4} & \textbf{60.0} & \textbf{73.4} & \textbf{87.9}  \\ 
			\bottomrule
	\end{tabular}}
	\label{tab:retrieval_voc}
\end{table}

\subsection{Results of Action Retrieval}
% \noindent \textbf{Comparison with the State-of-the-art.} 
% To maximize the retention of the original experimental setup, we did not evaluate the metrics when the Raranking Transformer method does not use reranking. We train these models and our proposed model on the Stanford-40 dataset and PASCAL VOC 2012 Action dataset respectively, and compare them with ours.

Table~\ref{tab:retrieval_stanford} shows the results of action retrieval on the Stanford-40 dataset. Our approach outperforms the baseline methods across all metrics. Specifically, without reranking, our method surpasses the R50-Token \cite{wu2022learning} by 10.3\% in terms of mAP and 4.4\% in Rank-1. Furthermore, when reranking is employed, our approach beats the R50-Token \cite{wu2022learning} by 5.7\% and 3.7\% for the mAP and Rank-1, respectively. The Reid strong baseline \cite{luo2019bag} is unsuitable for action retrieval, exhibiting a 14.1\% lower performance than ours in terms of mAP when reranking is not used.

The results on the PASCAL VOC dataset are summarized in Table~\ref{tab:retrieval_voc}. Similar to results on the Stanford-40 dataset, the worst-performing method is no longer DELG \cite{cao2020unifying}, followed by the Reid Strong Baseline \cite{luo2019bag}. Specifically, without reranking, our method outperforms the R50-Token \cite{wu2022learning} by 10.0\% in terms of mAP and 7.6\% in Rank-1. Additionally, when reranking is applied, our approach surpasses the R50-Token \cite{wu2022learning} by 7.8\% for mAP and 10.5\% for Rank-1. These experimental results confirm the effectiveness of our method for image-based action retrieval.

\subsection{Results of Action Recognition}
As our approach is trained utilizing the classification loss, we also present the results of image-based action recognition. Consistent with prior works \cite{zhao2017single, wu2020part}, we report the mean Average Precision (mAP) metric for action classification. We have chosen four representative methods: Action Mask \cite{zhang2016action}, Single Image Action Recognition \cite{zhao2017single}, the fine-tuned FasterRCNN \cite{bas2022top}, and ResNet-50 \cite{he2016deep}. The baseline single image action recognition \cite{zhao2017single} uses features of the whole image and the human box to predict the action in the image. 
% The method Action Mask \cite{zhang2016action} accurately delineates the foreground regions of underlying human-object interactions and reaches 82.6\%.  The Baseline network of Single image action recognition \cite{zhao2017single} uses features in the whole and the human box to predict the action in the image. 
% The FasterRCNN \cite{ren2015faster} is an object detection method that can detect and classify the object in the image. We use the finetuned FasterRCNN to recognize the actions (the reported result comes from \cite{bas2022top}). 

Table~\ref{tab:cls_s40} shows results of action recognition on the Stanford-40 dataset. The widely used ResNet-50~\cite{he2016deep} only has an mAP of 81.2\%. The part-based method Action Mask~\cite{zhang2016action} and the baseline~\cite{zhao2017single} achieve the mAP of 82.6\% and 84.2\%, respectively. Our proposed achieves 89.0\% mAP on action classification, and outperforms the baseline by 4.8\%.

\begin{table}[tbp]
	\caption{Results of image-based action recognition on the Stanford-40. \RED{The symbol ``*'' signifies human pose-based methods that utilize human pose as auxiliary information, whereas the alternative approach solely necessitates the global human bounding box.}  }
	\centering 
	\resizebox{0.45\textwidth}{!}{ 
		\begin{tabular}{l|c}
			\toprule
			Method & mAP \\
			\midrule
			ResNet-50 \cite{he2016deep} & 81.2 \\
			Action Mask \cite{zhang2016action} & 82.6 \\
			Single image action recognition (Baseline) \cite{zhao2017single} & 84.2\\
			\RED{Body Structure Cues~\cite{li2020recognizing}\textsuperscript{*} } & 93.8 \\
			\RED{Swin-T} \cite{liu2021swin}  & 91.1 \\
			FasterRCNN finetune \cite{bas2022top} & 78.0 \\
			\RED{MASPP \cite{ashrafi2023still}\textsuperscript{*} } & 94.8 \\
			\midrule
			\RED{RIART (ResNet-50)} & 89.0 \\
			\RED{RIART (Swin-T)} & \textbf{91.8} \\
			\bottomrule
	\end{tabular}}
	\label{tab:cls_s40}
\end{table}

\subsection{Ablation Studies}
We perform ablation experiments to assess the effectiveness of each component of our approach. First, we validate the three aspects of representations. Subsequently, we investigate the significance of the fusion transformer module. Furthermore, we evaluate the impact of the number of Transformer blocks in the fusion transformer module. Lastly, we evaluate the efficacy of positional and segment embedding. The ablation studies are conducted on the Stanford-40 dataset. 
% We undertake ablation studies on the Stanford-40 dataset and present the metrics obtained after reranking.

\begin{figure}[tbp]
	\subfigure[]{
		\label{fig:param_analysis:a}
		\includegraphics[width=1.58in]{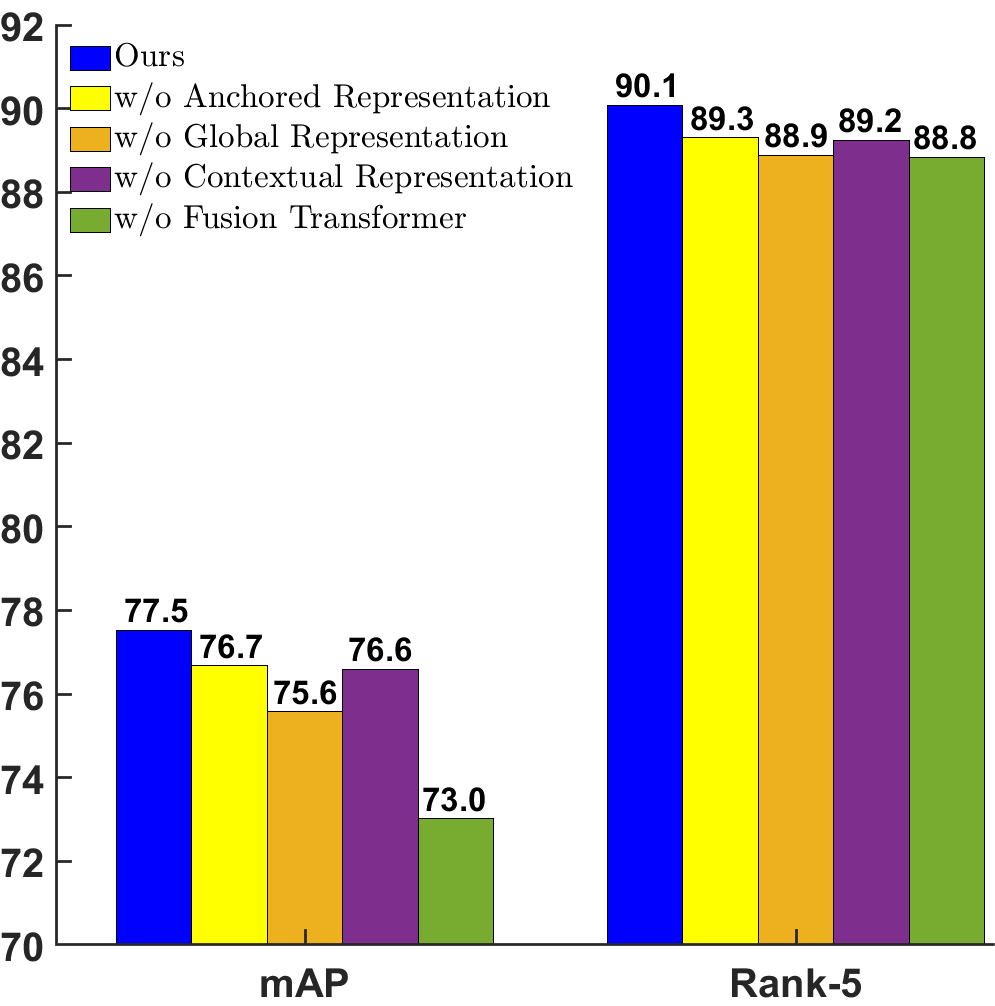}}
	\hspace{0.00in}
	\subfigure[]{
		\label{fig:param_analysis:b}
		\includegraphics[width=1.58in]{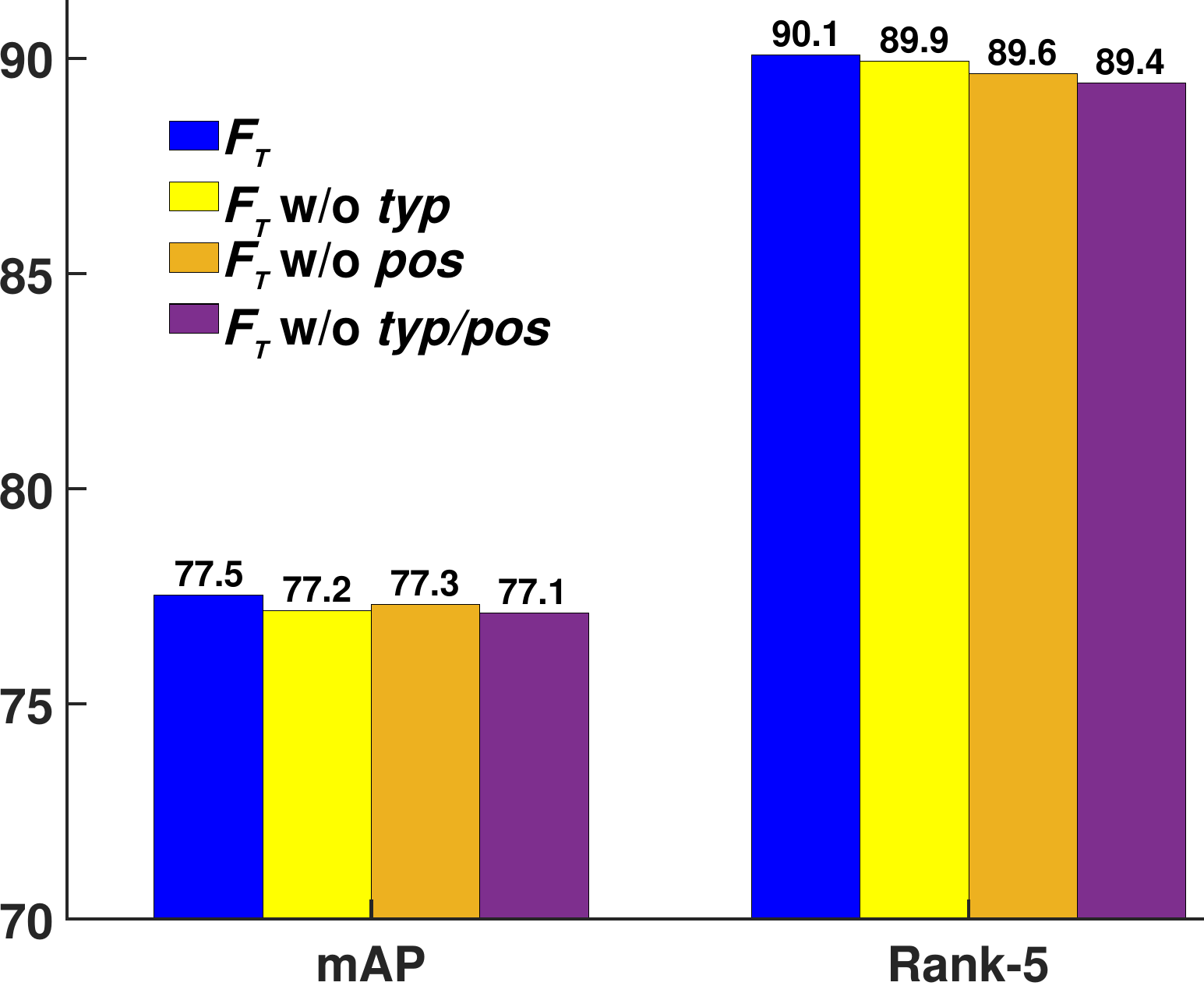}}
	\caption{(a) Ablation studies of features. (b) Ablation studies of embeddings in the fusion transformer.}
	% module on the Stanford-40 dataset
\end{figure}

\noindent \textbf{Multi-Level Representations.} There are three aspects of features: the anchored representation $F_{a}$, the global representation $F_{g}$, and contextual representations $\mathbf{F_{l}}$. 
Fig.~\ref{fig:param_analysis:a} shows the results of using different feature combinations on the Stanford-40 dataset. Using all three kinds of features, our approach achieves the best performance in terms of mAP and rank-5 accuracy compared to any combination of two kinds of features. When removing the $F_{a}$, the mAP decreases by nearly 0.8\%. Similarly, the mAP falls by approximately 1.9\% upon removing $F_{g}$.  Furthermore, our approach outperforms the variant that excludes the feature $\mathbf{F_{l}}$ by 0.9\%. This ablation study demonstrates that all three aspects of features contribute significantly to learning an effective representation of the action image.

\noindent \textbf{The Fusion Transformer Module.} We remove the fusion transformer module and instead use the average value of the input token sequences as a representation of the action image. The results on the Stanford-40 dataset are also shown in Fig.~\ref{fig:param_analysis:a}. We observe that the mAP and rank-5 accuracy decline significantly by 4.5\% and 1.3\%, respectively, upon removing the fusion transformer module. This demonstrates that the fusion transformer effectively merges the features into a superior representation of the action image.

\noindent \textbf{The Positional and Type Embedding.} We also investigate the significance of positional embedding $pos$ and type embedding $typ$ utilized in our model. Fig.~\ref{fig:param_analysis:b} presents the results obtained using different embeddings on the Stanford-40 dataset. Our model attains optimal performance when utilizing both embeddings. Conversely, excluding both results in the poorest performance, leading to a decrease of 0.4\% in mAP and 0.7\% in rank-5. Evidently, the final representation benefits from both positional and type embeddings.

% results of original resnet
\begin{table}[tbp]
	\caption{Analyses of different number of blocks in the fusion transformer.}
	\centering 
	\begin{tabular}{l|ccc}
		\toprule
		No. of Blocks & mAP & Rank-1 & Rank-5\\
		\midrule
		N=1 & 76.6 & 81.0 & 89.6\\ 
		N=2 & 76.8 & 80.8 & 89.2\\
		N=3 & \textbf{77.5} & \textbf{81.5} & \textbf{90.1}\\
		N=4 & 77.0 & 81.2 & 89.3\\
		N=6 & 76.6 & 80.6 & 89.4\\
		\bottomrule
	\end{tabular}
	\label{tab:ablation_num_block_s40}
\end{table}
\begin{figure*}%[htbp]
	\includegraphics[width=0.95\linewidth]{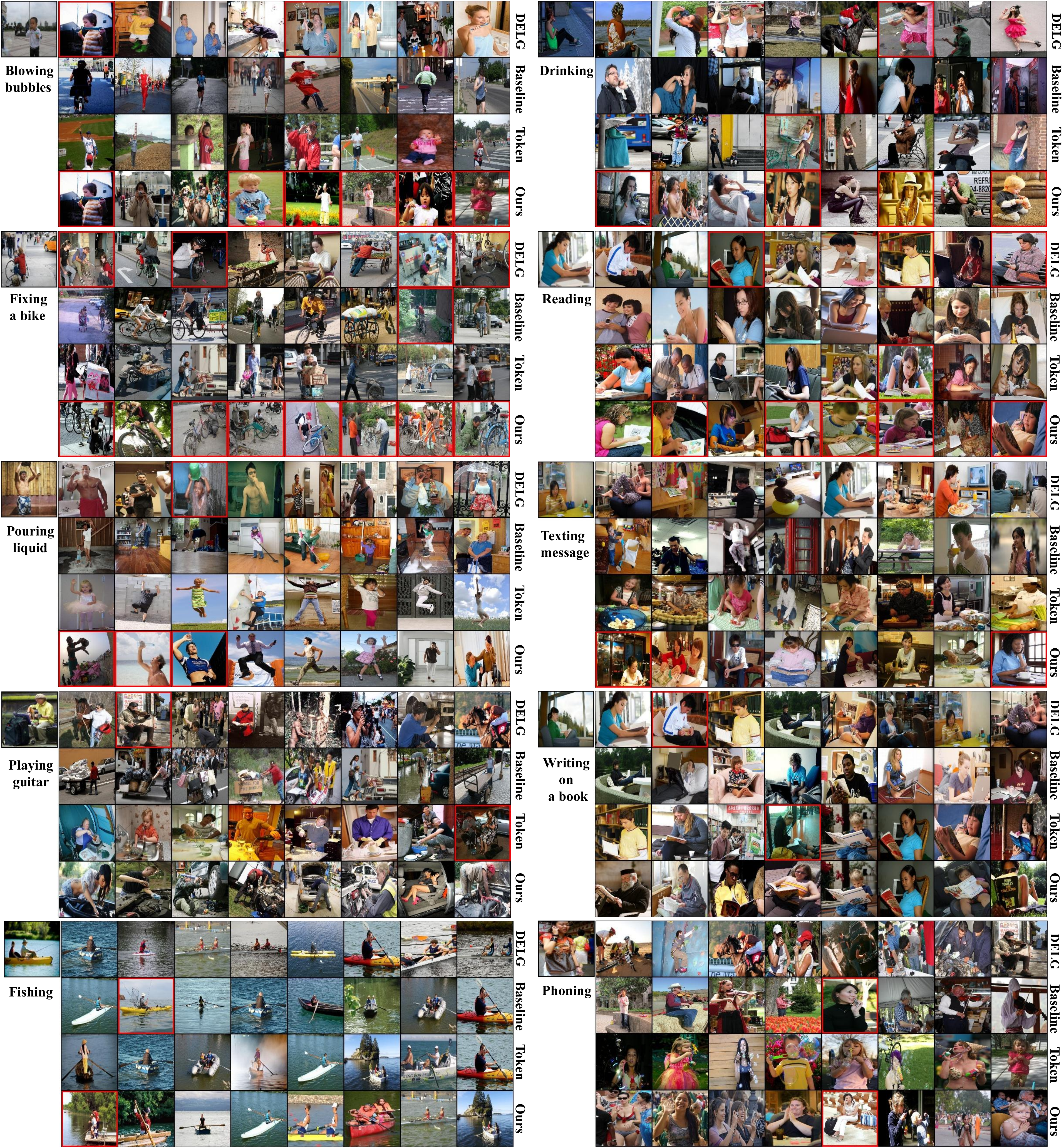}
	\caption{Visualization of retrieval results. For each example, the query image is on the left, and the top-$8$ retrieved images are displayed from left to right. \RED{The red box indicates the correctly retrieved sample, whereas the other images that do not have red boxes represent wrongly retrieved samples.} DELG, Baseline, and Token denote DELG global \cite{cao2020unifying}, Strong baseline \cite{luo2019bag}, Token \cite{wu2022learning}, respectively, which are representative instance-level image retrieval methods. The anchored human box in the query image is omitted for simplicity. \RED{The first four, i.e., ``blowing bubbles'', ``drinking'', ``fixing a bike'' and ``reading'', and our retrieved results are mostly accurate. The middle two, i.e., ``pouring liquid'' and ``texting message'' pose a bit of a challenge, and our approach can only correctly retrieve two out of the top-8 results. The last four, i.e., ``playing guitar'', ``writing on a book'', ``fishing'' and ``phoning'', are hard examples, and most approaches almost fail.}
	}
	\label{fig:vis_retrieval}
\end{figure*}

\noindent \textbf{The Number of Transformer Blocks.} To investigate the impact of varying the number of Transformer blocks, we experimented with settings of 1, 2, 3, 4, and 6 blocks, respectively. As shown in Table~\ref{tab:ablation_num_block_s40}, the performance initially improves and then deteriorates as the number of blocks increases. Having either too many or too few blocks can adversely affect performance. Our model attains the best performance when using three blocks.
% This may be due to the fact that the sample size of the dataset we used was not large enough, which led to overfitting when the number of parameters increased.

\subsection{Visualizations} 
Fig.~\ref{fig:vis_retrieval} shows the retrieval results for different methods. The image on the left is the query, and the red boxes note the correct retrieved samples from the gallery. Our proposed method outperforms other methods in terms of retrieval performance. For the action ``repair bicycle'', our method correctly retrieves 7 out of the top-8 results. In a challenging case for the action ``Drinking'', our method retrieves only 4 correct samples out of the top-8 results. Our method tends to misclassify some error samples as ``smoking'' actions, as they share significant similarities with the query action. \RED{The last four, i.e., ``playing guitar'', ``writing on a book'', ``fishing'' and ``phoning'', represent challenging examples for image-based action retrieval, and most approaches only achieve unsuccessful retrieval results.}
Nevertheless, in comparison to other methods, the error samples retrieved by our method are semantically more reasonable.

\section{Conclusion}  \label{sec:conclu}
\RED{In this paper, we empirically investigate the overlooked task of image-based human action retrieval, and establish new benchmarks and important baselines for this task. We introduce an efficient Region-aware Image-based human Action Retrieval with Transformers (RIART), which learns multi-level action representations from three perspectives: the anchored person, contextual regions, and the global image. A Transformer-based fusion module is also proposed to integrate these features into an enhanced representation.} Our approach achieves state-of-the-art performance on both action recognition and retrieval tasks. Furthermore, it demonstrates the capability to retrieve images depicting the same or similar actions as a given query image using a single image. Ablation studies highlight the effectiveness and complementarity of the three types of representations. We hope that this work and the established benchmarks can promote future research on image-based action retrieval.
% A limitation of our method is that it requires the human bounding box of the query image to filter the proposals. We plan to address this issue and further improve the performance of static action understanding in the future.
% \item Two authors: both authors' names and the year of publication;
% \item Three or more authors: first author's name followed by `et al.'

\section*{Acknowledgments}
This work is supported by the National Natural Science Foundation of China (62302093, 62172090, 62202438), Jiangsu Province Natural Science Fund (BK20230833), CAAI-Huawei MindSpore Open Fund, and the Southeast University Start-Up Grant for New Faculty (RF1028623063, RF1028623097). This work is also supported by the Big Data Computing Center of Southeast University.

\bibliographystyle{model2-names}
\bibliography{refs}

\end{document}